\documentclass[review]{elsarticle}

\usepackage{color}
\usepackage{hyperref}
\usepackage{amsmath, bm}
\usepackage{subfigure}
\usepackage[figuresright]{rotating}
\usepackage{multirow}
\usepackage{color}
\graphicspath{{figs/}}

\journal{Journal of \LaTeX\ Templates}

\begin{document}

\begin{frontmatter}

\title{Canonical Mean Filter for Almost Zero-Shot Multi-Task classification}

\author{Yong Li}

\author{Heng Wang}

\author{Xiang Ye\corref{mycorrespondingauthor}}
\cortext[mycorrespondingauthor]{Corresponding author}
\ead{yli@bupt.edu.cn}

\address{Beijing University of Posts and Telecommunications}

\begin{abstract}

The support set is a key to provide conditional prior for fast adaption of model in few-shot tasks.
But the strict form of support set makes its construction actually difficult in practical application.
Motivated by ANIL, we rethink the role of adaption in the feature extractor of CNAPs, which is a state-of-the-art 
representative few-shot method. To investigate the role, 
Almost Zero-Shot (AZS) task is designed by fixing the support set to replace the common scheme, which
provides corresponding support sets for the different conditional prior of different tasks.
The AZS experiment results infer that the adaptation works little in the feature extractor. 
However, CNAPs cannot be robust to randomly selected support sets and perform poorly on some datasets of Meta-Dataset
because of its scattered mean embeddings responded by the simple mean operator.
To enhance the robustness of CNAPs, Canonical Mean Filter (CMF) module is proposed to 
make the mean embeddings intensive and stable in feature space by mapping the support sets into a canonical form.
CMFs make CNAPs robust to any fixed support sets even if they are random matrices.
This attribution makes CNAPs be able to remove the mean encoder and 
the parameter adaptation network at the test stage, 
while CNAP-CMF on AZS tasks keeps the performance with one-shot tasks. 
It leads to a big parameter reduction. Precisely, 40.48\% parameters are dropped at the test stage.
Also, CNAP-CMF outperforms CNAPs in one-shot tasks because 
it addresses inner-task unstable performance problems. 
Classification performance, visualized and clustering results verify that CMFs make CNAPs better and simpler.

\end{abstract}


\end{frontmatter}

\section{Introduction}

A critical step in few-shot learning is to conduct the adaptation with the prior information provided by 
the support sets \cite{NIPS2016MatchingNet}. 
Ideally, a support set ought to contain some samples with their corresponding labels for each category \cite{NIPS2016MatchingNet}. 
Existing works often construct a support set for every particular task 
which deals with only a limited and known number of categories. 
In real few-shot applications however, it is almost impossible to construct a proper support set for 
the whole test dataset due to the unknown number of object categories and their labels in it. 
Since the prior information provided by an improper support set may misguide the adaptation, 
this work examines the role of the feature adaptation and the generalization ability of the trained 
model when proper support sets are unavailable. 

In literature, meta-learning methods mainly include two categories, 
the optimization-based and the metric-based methods. 
For optimization-based MAML (Model-agnostic Meta-learning) \cite{finn17maml}, ANIL (Almost No Inner Loop) \cite{Raghu2020Rapid}
is proposed to investigate the role of the adaptation 
of feature extractor by stopping the inner loop of MAML at the test stage. 
ANIL finds that the feature adaptation contributed
much less than the feature reuse in the generalization ability of MAML-based models. 
Motivated by ANIL, this work studies the role of the adaptation of feature extractor in 
metric-based few-shot methods. 
Among the metric-based methods are the representative CNAPs \cite{requeima2019cnaps}. 
CNAPs are evaluated on a large-scale and complex dataset containing 10 cross-domain common datasets while 
other metric-based methods mainly focus on in-domain transfer ability. 
Since the good transfer ability of CNAPs, this work employs them to study the contribution of the feature adaptation and feature reuse.

\begin{figure}[t]
	\centering
	\subfigure[Clustering results when the model is trained on Omniglot]{
		\includegraphics[scale=0.35]{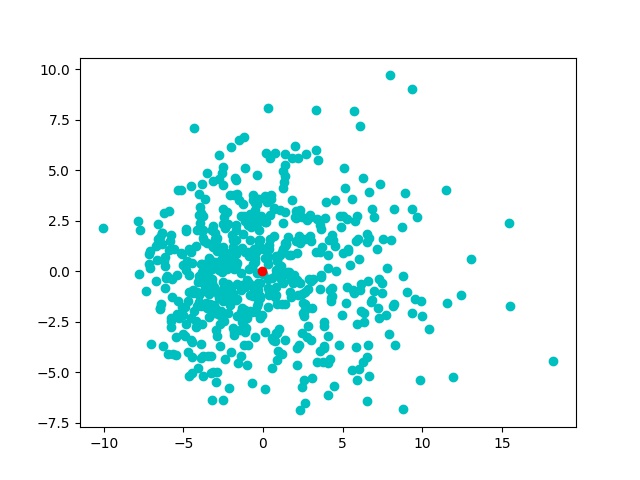}
		\label{subfigs:mean_clustering_omniglot} 
	}
	\subfigure[Clustering when the model is trained on Meta-Dataset]{
		\includegraphics[scale=0.35]{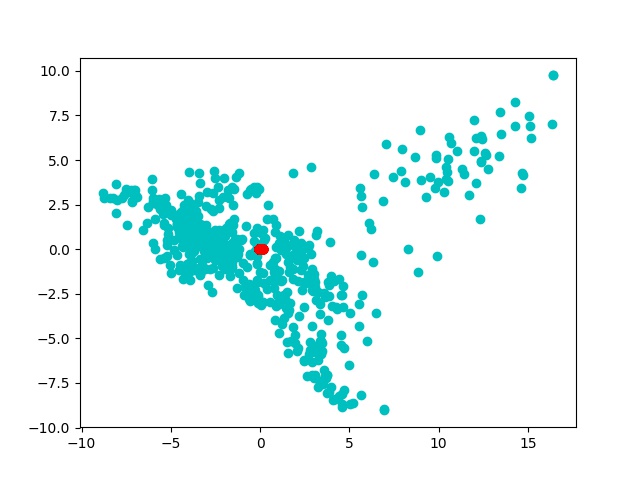}
		\label{subfigs:omniglot_clustering_trained_MD}
	}
	\caption{The clustering results of the extracted mean features 
		on Omniglot test images. }
	\label{figs:mean_clustering}
\end{figure}

Following the methodology of ANIL, this paper designs two `Almost Zero-Shot' (AZS) experiments, 
AZS-\uppercase\expandafter{\romannumeral1} and AZS-\uppercase\expandafter{\romannumeral2}, to analyze
the role of feature adaptation in CNAPs. The test Meta-Dataset \cite{Eleni2020MetaDataset}
containing 10 sub-datasets is employed. In AZS-\uppercase\expandafter{\romannumeral1}, 
one single random but fixed support set is chosen from each sub-dataset for all tasks in it to stop the adaptation, 
Only slight performance drop was observed. 

In AZS-\uppercase\expandafter{\romannumeral2}, a single random but fixed support set 
is chosen from a particular sub-dataset but 
is used across all sub-datasets to further stop all the adaptation for CNAPs 
at the inference stage. Again, only slight performance drop is observed 
on seven sub-datasets (see Table \ref{tab:cnaps_zeroshot_MD}) when the fixed 
support set is randomly selected from ImageNet.
This observation coincides with \cite{Raghu2020Rapid} in that task-specific support sets
are not necessitated for the feature extractors of few-shot learning methods.  
However, a dramatic performance drops was observed on few sub-datasets: 
Omniglot \cite{Science2015Omniglot}, Quick Draw, and Aircraft sub-datasets. 
Thus, a question arises: are the task-specific support sets still necessary for multi-dataset 
few-shot learning or the feature extractor of CNAPs relies too much on the support sets?

By deeply analyzing CNAPs, we find that the feature extractor is over-sensitive 
to the input support set, i.e., a small change of the support set may cause a dramatic 
adaptation shift when guiding the inference model. Fig. \ref{figs:mean_clustering}
shows an example of scattered mean features of corresponding support sets for different tasks. 
Consequently, a support set taken from a particular sub-dataset may perform poor across sub-datasets. 

To address the over-sensitivity, this work proposed a dynamic-kernel-like module, Canonical Mean Filter (CMF)
\footnote{CMF is a lightweight module which can be embedded in the mean encoder of original CNAPs, called CNAP-CMF.}. 
CMF firstly applies a designed attention module to generate a vector for each sample, and then 
fuses the attention vectors over all image samples belonging to a support set. Then the fused 
vector serving as the set of weights are assigned to the kernel in next layer. 
In CMF, the kernels are able to map varying samples to a canonical form and  
become less sensitive to the input samples,  so the mean prior provided by CMF is stable 
even when the support sets are randomly generated. 
Fig. \ref{subfigs:mean_clustering_omniglot} and \ref{subfigs:omniglot_clustering_trained_MD} 
show the intensive and stable mean features (red points) extracted by CMF.

Similar to ANIL \cite{Raghu2020Rapid}, the proposed CMF provides a better feature reuse. 
To investigate the contribution of feature reuse to the classification performance, 
we conducted various experiments 
on both AZS-\uppercase\expandafter{\romannumeral2} and one-shot tasks. 
Since the classifiers require the support sets \cite{Raghu2020Rapid}, which means the adaptation still 
contributes to the classification performance, to separate the feature reuse from the adaptation, 
we resort to analyzing the clusterability of the features extracted by CNAP-CMF.  The analysis
shows that they have a much smaller inner-class distance than CNAPs-extracted features
while maintaining a comparable inter-class distance 
(see detail in Section \ref{subsec:clustering_analyses}), indicating a better clusterability. 

As a result, the feature adaptation can be stopped in CNAP-CMF, i.e., 
the adaptation parameters can be pre-computed and stored before the test stage to 
avoid repetitive computation.
Then the stored adaptation parameters can be reused at the test stage.
So the mean feature encoder and parameter adaptation networks are removed in the test stage, 
leading to the reduction of parameter amount by 40.48\%.
The main contributions are concluded as follows.

\begin{enumerate}
	\item 
	Motivated by ANIL, we show that the feature reuse is more important
	in metric-based few-shot methods by experimenting with Almost Zero-Shot (AZS) tasks, AZS-I and AZS-II.
	In AZS-II tasks, the adaptation can be completely stopped, which only needs a 
	randomly chosen support set from any arbitrary sub-dataset in the test stage. 
	
	\item  
	CMF is proposed to address the problem of scattered mean embeddings yielded by CNAPs
	by mapping varying support sets to a canonical form. 
	The canonical form means stable mean embeddings, enabling CNAP-CMF to outperform CNAPs 
	by a large margin on AZS-II and one-shot tasks.

	\item 
	After CNAP-CMF is fully trained, the adaptation of the feature extractor is stopped.
    Naturally, the structures responsible for feature adaptation can be removed in the test stage,
	and only the frozen ResNet-18 and the stored adaptation network of linear classifiers need to be stored. 
	This reduces the number of parameters by 40.48\%.
	 
\end{enumerate}

The rest of the paper is organized as follows. 
Section \ref{sec:related_works} overviews the meta-learning and few-shot learning methods; 
Section \ref{sec:methods} formulates CNAPs and CNPs \cite{Garnelo2020CNP}  
in a mathematical form, then presents the canonical mean filter (CMF); 
Section \ref{sec:experiments} presents the experimental results on multi-task classification, 
and conducts the ablation experiments to analyze the proposed CMF; and 
the conclusion is given in Section \ref{sec:conclusion}.

\section{Related Works}
\label{sec:related_works}

{\color{blue}
Few-shot learning methods can be divided into metric-based methods and gradient-based methods. 
For metric-based methods, Authors \cite{NIPS2016MatchingNet} 
proposed the episode fashion for deep few-shot learning and 
designed the MatchingNets addressing few-shot tasks.
Following \cite{NIPS2016MatchingNet}, 
Snell \textit{et al.} \cite{Snell2017ProtoNets} proposed Prototypical Networks for Few-shot Learning, 
which used a KNN-like method to determine the unseen class of a test sample
according to the feature distance between 
the test sample and each support sample. 

For gradient-based methods, MAML \cite{finn17maml} was proposed to yield adaptive gradients
in inner loop for fast adaptation in test stage. 
Following \cite{finn17maml}, many variants of MAML were designed. 
FOMAML \cite{Biswas2018FirstorderMI} dropped the second-order item to avoid computing the Hessian matrix in MAML.
Nichol \textit{et al.} proposed Reptile \cite{Nichol2018OnFM} to address 
the problem of computing the complex Hessian matrix  
in MAML by an analytical first-order gradient.
Wang \textit{et. al.} proposed Meta-Metric-Learner \cite{WANG2019202} to combine 
the advantages of both the metric-based and optimization-based 
methods for handling the varying number of classes, 
and hence generate more generalized metrics for classification across tasks.

Since Conditional Neural Processes \cite{Garnelo2020CNP} was proposed, 
the CNP-based methods proposed in \cite{Garnelo2020CNP,garnelo2018neural,kim2019attentive,gordon2019convolutional,foong2020convLNP}
have become more and more popular because of their simplified computation of complex covariance matrix in 
KNN-based methods and their satisfying performance on 1-D and 2-D regression tasks. 
NPs \cite{garnelo2018neural} futher added real random sample in CNPs and used the Reparameterization Trick proposed in Variational
Autoencoder \cite{kingma2013vae}. 
ANPs \cite{kim2019attentive} incorporated the attention mechanism commonly used in the Natural Language Process (NLP), 
which includes the single-head and multi-head attention, to make the conditional inference of NP be more focused on the key information.
ConvCNPs \cite{gordon2019convolutional} make the CNPs capable of modeling translation equivariance, 
which is an important inductive bias for many learning problems, in the data. 
CNAPs \cite{requeima2019cnaps} use a fully trained model to provide enough discrimination for the 
multi-task classification tasks, which improved the classification capability over the CNPs. 


CNP-based methods explicitly employ an adaptation module to adapt the parameters of 
feature-extractor structure to the unseen test samples. 
The ANIL \cite{Raghu2020Rapid} however, thought that 
the universal feature representation played a more important role than the adaptation module
for few-shot leaning. 
Following the ANIL, we further designed two experiments, AZS-I and AZS-II, and the 
experimental results coincide with the point view of the ANIL. 
Observing this, this work designed an attention module to reallize the 
canonical mean filter (CMF). The CMF provides a stable transfer feature that can guide
the feature extractor to generate the universal feature representation. 
}

{\color{magenta}
Similar to Dynamic Convolution (DY) \cite{2020DynamicConv}, 
the proposed CMF can be taken as a type of dynamic filters. But 
different from DY, the CMF 
aims to generate sample-independent features that  
are stable across different tasks;   
DY aims to generate sample-dependent features for the classification task
that are often (desired to be) significantly distinct across different classes. 
}

\section{Proposed method}
\label{sec:methods}
This section presents the canonical mean filter (CMF) that is able to map varying 
support sets to a canonical form. To clearly state the rationale of CMF, we firstly 
need to mathematically formulate the CNAP-based methods. 
With the formulation, the mean shifting is analyzed and shown to 
be over-sensitive to the variation of support sets. 
Then, we propose CMF to alleviate the over-sensitiveness by mapping the input images to a canonical form, 
and provide a theoretical explanation from the Bayesian inference view.

\subsection{Mathematical Formulation of CNAPs}
\label{subsec:form_and_analysis}

CNAPs originate from CNPs. For the completeness, the CNPs \cite{Garnelo2020CNP} are briefly introduced here. 
Let $\mathcal{X}^C$ and $\mathcal{X}^T$ denote context image dataset and target image dataset, 
and $\mathcal{Y}^C$ and $\mathcal{Y}^T$ denote their labels. 

CNPs use Neural Networks (NNs) to build the mean mapping parameterized by $\theta$
from the context image dataset to feature space, 
$F(\cdot, \theta):\mathcal{X}^{C} \rightarrow \mathcal{R}^{C}$. 
$\mathcal{R}^{C} = F(\mathcal{X}^{C}, \theta)$ encodes the distribution information of 
support sets $\{\mathcal{X}^C, \mathcal{Y}^C\}$ and 
is used to generate the adaptation parameter $\phi$ for the inference model by
\begin{displaymath}
\bm{\pi} (\cdot, \mathcal{R}^C) : \varphi \rightarrow \phi, 
\end{displaymath}
where $\varphi$ represents the original parameter of the inference model 
to be adapted.
We call the parameter adaptation \textit{mean shifting} in this work.
Then, the inference model $G(\phi, \cdot)$ predicts the classes of 
the target set $\mathcal{X}^T$ by 
\begin{equation}
\label{eq:cnp_ori_infer_model}
\mathcal{Y}_{pred}^T = G \left( \phi, \mathcal{X}^T \right) = G \bigg(\bm{\pi} \left(\varphi, F(\mathcal{X}^{C}, \theta) \right), \mathcal{X}^T \bigg). 
\end{equation}

In the training process, CNPs optimize $G$ jointly over $\theta$ and $\varphi$ in Equation \ref{eq:cnp_ori_infer_model}
by maximizing the likelihood given $(\mathcal{X}^T, \mathcal{Y}^T)$. Formally, 
\begin{equation}
\label{eq:cnp_opt_likelihood}
\hat{\theta}, \hat{\varphi} = \mathop{\arg\max}_{\theta, \varphi} \ \ p \left(\mathcal{Y}^T | G \left(\bm{\pi} \left(\varphi, F(\mathcal{X}^{C}, \theta) \right) , \mathcal{X}^T \right ) \right) .
\end{equation}

CNAPs \cite{requeima2019cnaps} build a simple convolutional neural network to get $\mathcal{R}^C$, 
then use a mean operator to aggregate the features over all samples to get the mean vector $\mathcal{R}^C_m$, 
$$\mathcal{R}^C_m = \frac{1}{S} \sum_{i = 1}^{S} \mathcal{R}^C_i,$$
where $S$ is the candinality of the input support sets.

Following the FiLM layers \cite{perez2018film}, CNAPs design 
an adaptation network parameterized by $\bm \gamma = \{\gamma_w,\gamma_b\}$ to 
realize the mapping $\bm{\pi}$ in Eq. \ref{eq:cnp_ori_infer_model}.
Formally, 
\begin{equation}
\label{eq:adaption_resnet}
\phi = \bm{\pi} (\varphi,\bm \gamma, \mathcal{R}^{C}_m) = (\gamma_w \otimes \mathcal{R}^{C}_m) \times \varphi + (\gamma_b \otimes \mathcal{R}^{C}_m), 
\end{equation}
where $\otimes$ represents the convolution operation.

To build a powerful feature extractor for the inference model, CNAPs use the ResNet-18 \cite{He2015Deep} 
with every convolutional layer being followed by FiLM layers \cite{perez2018film}.
The ResNet-18 is fully trained on ImageNet-1K and outputs 512-D embeddings for the final linear classification.
After being trained on ImageNet-1K, 
the parameters $\varphi$ of the ResNet18 are frozen, i.e., the they will not be updated
in the subsequent training process of CNAPs. 
$\bm{\pi} (\varphi,\bm \gamma, \mathcal{R}^{C}_m)$ adapts $\varphi$ of the frozen fully-trained ResNet-18 model to 
a new target dataset 
according to $\mathcal{R}^{C}_m$ and $\bm \gamma$. 
With $\bm{\pi}$, the maximum likelihood in Equation \ref{eq:cnp_opt_likelihood} will be cast as 
\begin{equation}
\hat{\theta}, \hat{\bm \gamma} = \mathop{\arg\max}_{\theta, \bm \gamma} \ \ p \left(\mathcal{Y}^T \big| G \left(\bm{\pi} \left(\varphi, \bm \gamma, F(\mathcal{X}^C, \theta)\right), \mathcal{X}^T\right)\right).
\label{eq:cnaps_optim}
\end{equation}
In Eq. \ref{eq:cnaps_optim}, CNAPs optimize $G(\phi, \mathcal{X}^T)$ jointly over $\theta$ and $\bm \gamma$. 
Since $\bm \gamma$ will also be driven by $\mathcal{R}^C_m$, i.e., its update depends on $\mathcal{R}^C_m$, 
the over-sensitivity comes from $F(\cdot, \theta)$.
Thus, to alleviate the over-sensitivity of CNAPs to the varying support sets,
a critical step is to design 
$F(\cdot, \theta)$ such that $\mathcal{R}^{C}_m = F(\mathcal{X}^C, \theta)$ does not change dramatically with $\mathcal{X}^C$.
Motivated by this observation, this work proposes the canonical mean filter 
that generate stable features for the adaptation to realize the \textit{mean shifting}.

\subsection{Canonical Mean Filter}
\label{subsec:CMF}

To strengthen the stability of $\mathcal{R}^C_m=F(\mathcal{X}_C, \theta)$, 
this work proposed Canonical Mean Filter (CMF).
CMF maps different samples to a canonical form, 
reducing the over-sensitivity of CNAPs to the wide variation of support sets, 
and even helping remove the demand on the support sets. 
Note, it is important that $\mathcal{R}^C_m$ represents the mean vector for adaptation
on the whole dataset rather than only on a particular task. 
This hence means that $\theta$ is desired to be optimized by the marginal likelihood 
of all tasks instead of particular tasks. 
In fact, task-specific likelihood is more difficult to be optimized than marginal likelihood 
of all tasks in few-shot tasks; and the former gives an inferior performance to the latter 
\cite{patacchiola2020bayesian,Robbins2020EMPIRICALBAYES}.

\begin{figure}[t]
	\centering
	\includegraphics[scale=0.6]{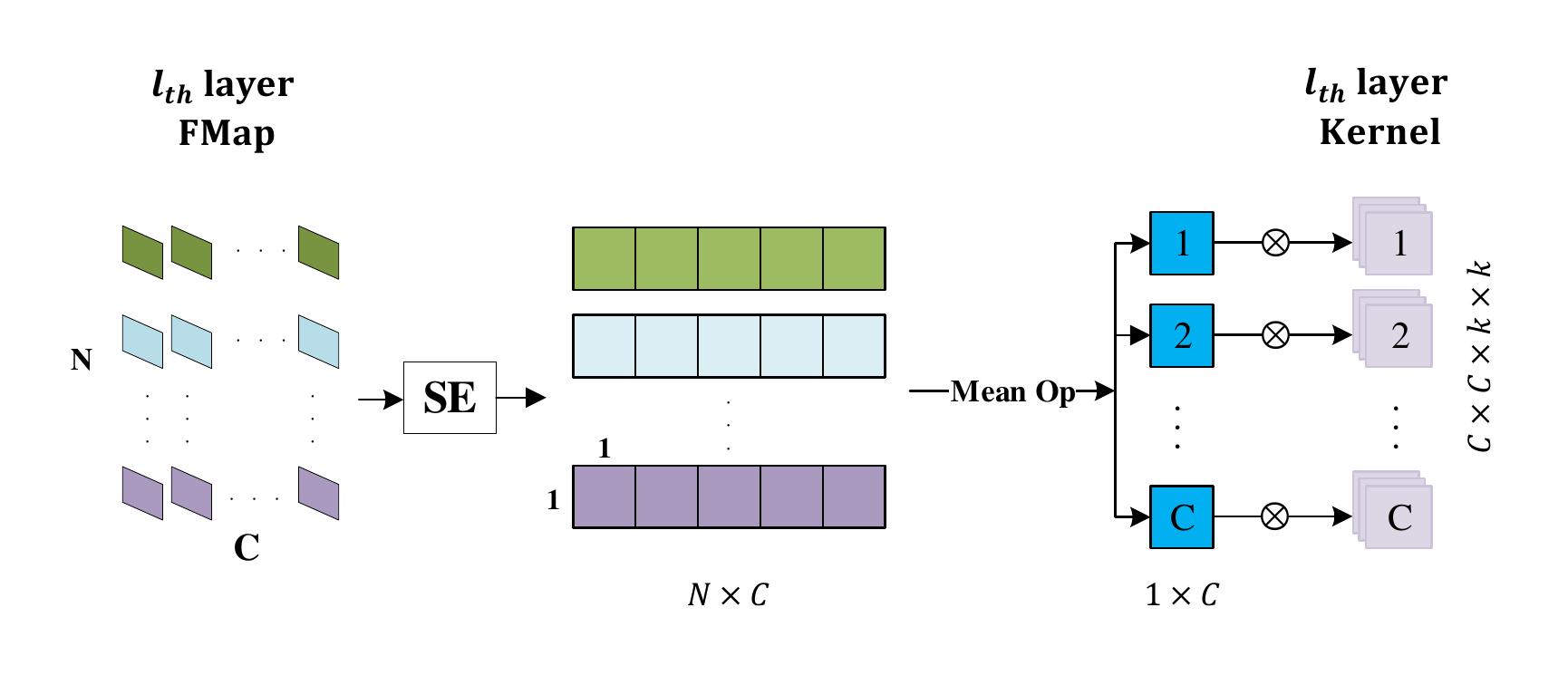}
	\caption{\textbf{Canonical Mean Filter architecture.} To build stable mean shifting, 
			the salient features of each support sample in the $(l-1)_{th}$ layer are extracted by Squeeze-and-Excitation module, 
			and then are averaged across all samples to get the canonical mapping weights for the kernels in the $l_{th}$ layer.}
	\label{fig:CMF_architecture}
\end{figure}

{\color{blue}
Seeing this, we design CMF to realize the marginal likelihood optimization over all tasks by a kernel way
to get representative and stable $\mathcal{R}^C_m$.
As is shown in Fig. \ref{fig:CMF_architecture}, a designed attention module is used to extract 
the most salient feature $\mathcal{SF}^{l-1} \in R^{N \times C}$ of each sample at the $(l-1)_{th}$ layer, 
and then $\mathcal{SF}^{l-1}$ are averaged across all samples for fusing features by 
\begin{equation}
\label{eq:salient_ftr_avg}
\overline{\mathcal{SF}}^{l-1} = \frac{1}{N} \sum_{0}^{N-1} \mathcal{SF}^{l-1}, 
\end{equation}
where $N$ is the number of samples in a task. $\overline{\mathcal{SF}}^{l-1}$ 
in Eq. \ref{eq:salient_ftr_avg} serving as the weights, are 
assigned to the kernels in the $l_{th}$ layer. 
Tab. \ref{tab:CMF_arch} gives the structure the designed attention module. 
}

\begin{table}[h]
	\centering
	\normalsize
	\caption{\color{blue} The architecture of the proposed CMF}
	\label{tab:CMF_arch}
	\begin{tabular}{ccc}
		operators  & kernel shape           & output shape \\
		\hline
		\hline
		Input      & $\mathcal{F}^{l-1}$    & $N \times C \times W \times H$ \\
		\hline
		Pooling    & Max Pooling            & $N \times C$ \\
		\hline
		FC1        & $C \times \frac{C}{4}$ & $N \times \frac{C}{4}$ \\
		\hline
		ReLU       & --                     & --               \\
		\hline
		FC2        & $\frac{C}{4} \times C$ & $N \times C$ \\
		\hline
		Output     & $\mathcal{SF}^{l-1}$   & $N \times C$ \\
	\end{tabular}
\end{table}

For explaining the CMF, we follow CNAPs and assume that all convolutional layers in $F(\cdot, \theta)$ contain the 
same number of kernels, and thus the output channel numbers are also same. Let $\mathcal{F}^l$ denote
the $l_{th}$ layer, $l \in \{1, \cdots, L\}$, and $\mathcal{K}^{l}$ denote the set of kernels in $\mathcal{F}^l$. 
CMF can be formulated as

\begin{equation}
\begin{split}
\mathcal{F}^l & = (\overline{\mathcal{SF}}^{l-1} \cdot \mathcal{K}^{l}) \otimes \mathcal{F}^{l-1} \\
              & = \overline{\mathcal{SF}}^{l-1} \cdot (\mathcal{K}^{l} \otimes \mathcal{F}^{l-1}) .
\end{split}
\label{eq:DMF}
\end{equation}

In Eq. \ref{eq:DMF}, $\overline{\mathcal{SF}}^{l-1}$ containing the average 
information of all samples will undermine the (biased) influenece of 
individual samples on $\mathcal{F}^l$. 
Like Dynamic Filter technology, 
the mean weights $\overline{\mathcal{SF}}^{l-1}$ assigned to the kernels in 
$\mathcal{F}^l$ will bring the model adaptability to unseen data \cite{2020Dynamic}.
In addition, 
the averaging and fusion are carried out at every convolutional layer in this work
rather than only at the final layer, which makes the fusion more 
effective in hierarchical architectures \cite{Robbins2020EMPIRICALBAYES} and hence 
the ideal mean embedding of $F(\mathcal{X}^C, \theta)$ more accessible.

{\color{magenta}
\subsection{Bayesian View of CMF}
\label{subsec:CMF_Bayes}
This section analyzes the CMF and CNAPs from Bayesian inference view. 
At the training stage, the mean encoder of the CNAPs were trained to maximize the 
the classification accuracy within a particular task.  This amounts to searching for the 
mean vector $\mathcal{R}^C_m$ that maximizes the likelihood 
of $p\left( \mathcal{Y}_k^T | \mathcal{X}_k^T, \mathcal{R}^C_m \right)$. Formally, 
\begin{equation}
\hat{ \theta } = \arg \max_{\theta}  p\left( \mathcal{Y}_k^T | \mathcal{X}_k^T, F(\mathcal{X}^{C}_k, \theta) \right), 
\label{eq:cnap_optimized_theta}
\end{equation}
where $\mathcal{R}^C_m = F(\mathcal{X}^{C}_k, \theta)$ and 
the support sets $\mathcal{X}^{C}_k$ are chosen from the task. 

The obtained $\hat{ \theta }$ by Eq. \eqref{eq:cnap_optimized_theta} is task-specific, and 
$F(\mathcal{X}^{C}_k, \hat{\theta})$ is expected to 
be able to yield a task-specific feature $\mathcal{R}^C_m$ that provides 
a good prior for parameter adaptation within the task. 
Unfortunately, 
the task-level optimization for $\theta$ does not perform satisfactorily 
on few-shot tasks \cite{patacchiola2020bayesian,Liu2020E3BM},
since 
$F(\mathcal{X}^{C}_j, \hat{\theta})$ will not provide a good prior
for an unseen task.

As a comparison, CMF can be formulated as a kernel method of searching
the Bayesian parameter $\theta$ that maximizes the marginal likelihood over all $K$ tasks. Formally, 
\begin{equation}
\label{eq:margin_CMF}
\begin{aligned}
\hat{ \theta } & = \arg \max_{\theta} \ \prod \limits_{k = 1}^K  p\left( \mathcal{Y}_k^T | \mathcal{X}_k^T, F(\mathcal{X}^{C}_k, \theta) \right)  \\
& = \arg \max_{\theta} \int \prod \limits_{k = 1}^K  p\left( \mathcal{Y}_k^T | \mathcal{X}_k^T, F(\mathcal{X}^{C}_k, \theta) \right) p(\mathcal{X}^{C}_k) d \mathcal{X}^{C}_k  \\ 
& = \arg \max_{\theta} \frac{1}{K}\int \prod \limits_{k = 1}^K  p\left( \mathcal{Y}_k^T | \mathcal{X}_k^T, F(\mathcal{X}^{C}_k, \theta) \right) d \mathcal{X}^{C}_k.
\end{aligned}
\end{equation}
where $p(\mathcal{X}^{C}_k)$ always equals to $\frac{1}{K}$, 
and $\{\mathcal{Y}^{T}_k, \mathcal{X}^{T}_k\}, k=1, \ldots, K$, denotes the target set for the  $k_{th}$ task. 
Once $\hat{ \theta } $ is  obtained from Eq. \eqref{eq:margin_CMF}, 
$\hat{\mathcal{R}}^C_m = F(\mathcal{X}^C, \hat{ \theta })$ with any arbitary $\mathcal{X}^C$
can provide the best average classification performance on all tasks since 
the influence of $\mathcal{X}^{C}$ on $\hat{\mathcal{R}}^C_m$ is significantly suppressed.
	
Patacchiola \textit{et al.} proposed Deep Kernel Transfer (DKT) \cite{patacchiola2020bayesian} 
to 6approximate the kernel parameterized by $\theta$ with 
the kernel of the Gaussians Process (GP) in Eq. \ref{eq:margin_CMF} and realized the marginal optimization.  
However, GP kernels cannot approximate many generally unknown kernels, which leads to a bad fitting model.
Although many kernels are generally unknown, we can use the output of the dynamic kernels to approximate the output of the unknown kernels.
The CMF designs a canonical filter to yield a stable mean prior, approximating
the prior generated by the $\hat{ \theta }$ resulting from the marginal optimization over all support sets in Eq. (8). 

}

\section{Experiments}
\label{sec:experiments}

    In this section, CNAP-CMF is compared with the state-of-the-art CNAPs for 
	multi-task classification tasks on Meta-Datasets. Three tasks 
	are designed in Section \ref{subsec:multitask_classification} to investigate the role of feature reuse:
	AZS-I, AZS-II, and one-shot tasks. Similar to CNAPs, the models are fully trained on 
	two different training datasets, 1) all sub-datasets of Meta-Datasets, 
	and 2) ImageNet-1K. 
		The trained models are then tested on 
	all sub-datasets of Meta-Datasets as well as MNIST, CIFAR-10, and CIFAR-100, 
	in order to evaluate the cross-domain performance. 		
	Then the clustering performance of CNAPs and CNAP-CMF are analyzed 
	in Section \ref{subsec:clustering_analyses}
	to investigate the linear separability.

\subsection{Multi-task Classification}
\label{subsec:multitask_classification}

This section reports the evaluation results of AZS-II and one-shot tasks 
on the whole test dataset of Meta-Dataset when the 
models are trained on Meta-Dataset.
Also, the evaluated results of CNAPs (AR) and CNAP-CMF (AR) for one-shot tasks, whose 
models are fully trained on ImageNet-1k, are also reported.
The result of AZS-I is reported in Table \ref{tab:AZS-I_experments}. 

\subsubsection{Training Strategy}

We follow the training strategy in CANPs \cite{requeima2019cnaps}, 
and only 2 NVIDIA Ti-1080 GPUs with 12G memory are used. 
Due to the limit of computation resource, 
the auto-regressive (AR) proposed in CNAPs is not used 
when the model is trained on Meta-Dataset, 
but used when the model is trained on ImageNet-1K.

All models are trained for 110,000 episodes with the Adam \cite{Kingma2014Adam} optimizer. 
The learning rate is 0.0005 and the batch size is 16. 
The trained model is validated per 200 episodes during the training process. 
For the multi-dataset classification task, 
a model is considered to perform better than current model if it 
yields higher classification accuracy on over half of all sub-datasets.

\subsubsection{Training on Meta-Dataset} 

Tab. \ref{tab:cnaps_zeroshot_MD} gives
the classification performance of CNAP-CMF on AZS-II tasks.  
On Omniglot dataset, AZS-II, CNAP-CMF outperforms CNAPs 37.2\%. 
Similarly, for AZS-II, CNAP-CMF outperforms CNAPs by 38.3\%, 20.7\%, 
and 10.4\% on Aircraft, Quick Draw, and MNIST datasets respectively.
When CNAPs are directly applied to AZS-II tasks, 
where the fixed support set is randomly chosen in ImageNet-1k sub-dataset, 
the classification performance decreases significantly on  
Omniglot \cite{Science2015Omniglot}, Aircraft \cite{maji13fine-grained}, Quick Draw, and MNIST. 
When the fixed support set is chosen in other sub-datasets, 
Table \ref{tab:each_support_selection} gives severer performance drop on all evaluated sub-datasets.
CNAP-CMF on AZS-II tasks performs only slightly
inferior to CNAPs on one-shot tasks. 

\begin{table}[htb]
	\centering
	\scriptsize
	\caption{Multi-task image classification results. CNAPs and CNAP-CMF are tested on 
		AZS-II and one-shot tasks respectively.
	}
	\label{tab:cnaps_zeroshot_MD}
	\begin{tabular}{l|cccc}
	     \hline
		\multirow{2}{*}{Datasets} & CNAP & CNAP & CMF & CMF \\
		& OneShot & AZS-II & OneShot & AZS-II \\
		\hline
		ImageNet & 51.3     & 50.8      & 51.6    & 51.2 \\
		Omniglot & 88       & 50.2       & 87.7    & 87.4 \\
		Aircraft & 76.8     & 39.1      & 77.9    & 77.4 \\
		Birds    & 71.4     & 69.3     & 71.6    & 69.9 \\
		Textures & 62.5     & 64        & 63.9    & 63.7 \\
		Quick Draw & 71.9   & 49.8      & 71.6    & 70.5 \\
		Fungi    & 46       & 40.4      & 45.4    & 43.4 \\
		VGG Flower & 89.2   & 80.5      & 89.7    & 89.4 \\
		Traffic Sign & 60.1 & 59.1      & 61.3    & 61.4 \\
		MSCOCO   & 42       & 41.9      & 44.1    & 43   \\ 
		MNIST    & 88.6     & 78.7      & 89.2    & 89.1 \\
		CIFAR10  & 60       & 58.7     & 66.1    & 66.3 \\
		CIFAR100 & 48.1     & 47.6      & 50      & 51.3   \\
		\hline
	\end{tabular}
\end{table}

For one-shot tasks, CNAP-CMF also outperforms CNAPs by $0.3\% \sim 6\%$ on most test sub-datasets.
The reason is that different support sets in 
the same task can yield classification performance fluctuation due to unstable mean embeddings. 
To simply see the fluctuation, four support sets are randomly chosen from Omniglot dataset in 
the same task, and Table \ref{tab:inner_class_difference_omniglot} gives the classification results. 
For example, on Task 3, the support set $\mathcal{X}^S_1$ gives 76.5\% classification accuracy, 
but $\mathcal{X}^S_4$ gives 92\%, 15.5\% higher than $\mathcal{X}^S_1$. As a comparison, 
CMF addresses the inner-task performance fluctuation by generating stable 
mean embeddings which are almost independent of the support sets in a particular task.

\begin{table}[htb]
	\centering
	\normalsize
	\caption{Classification result of CNAPs using different support sets for the same task on Omniglot. Unit: \%.}
	\label{tab:inner_class_difference_omniglot}
	\begin{tabular}{l|cccc}
		Task & $\mathcal{X}^S_1$ & $\mathcal{X}^S_2$ & $\mathcal{X}^S_3$ & $\mathcal{X}^S_4$ \\
		\hline
		T1                     & 93.5 & 88.9 & 91.5 & 90.1 \\
		T2                     & 81.0 & 90.0 & 85.0 & 85.5 \\
		T3                     & 76.5 & 80.0 & 79.5 & 92.0 \\
	\end{tabular}
\end{table}

\subsubsection{Training on Single Dataset}

Due to the limit of computation resource, the proposed AR in CNAPs 
that needs to be trained on the whole Meta-Dataset
is not used in the following evaluation experiments. 
For verifying the effectiveness of the proposed CMF and maintaining the 
thoroughness of the whole evaluation experiments, AR is used in the following comparison experiments, 
in which the models are only trained on ImageNet sub-dataset.

As is given in Tab. \ref{tab:comparison_zerooneshot_singleData}, CNAP-CMF (AR) outperforms
CNAPs (AR) on almost all sub-datasets of Meta-Datasets. 
Although the model is only trained on ImageNet-1K, the pre-trained ResNet-18 model brings the inference enough  
separability capability to make the CNAPs (AR) can be generalized to other datasets. 
Also, CMF making the mean mapping $F(\cdot, \theta)$
learn a better mean representation when the unseen data is input.
The better representation allows the inference model $G(\phi, \mathcal{X}^T)$ to be adapted to the target distributions
more correctly.

\begin{table}[h]
	\centering
	\scriptsize
	\caption{Classification results of CNAPs-One (AR), CNAP-CMF-One (AR) and other SOTA methods are compared.
		The evaluated models are trained on ImageNet-1K only, and evaluated on the whole Meta-Dataset.}
	\label{tab:comparison_zerooneshot_singleData}
	\begin{tabular}{l|cccc}
		\hline
		\multirow{2}{*}{Datasets} & ProtoNet & MatchingNet & CNAP & CMF \\
		& One (AR) & One (AR) & One (AR) & One (AR) \\
		\hline
		ImageNet & 50.5     & 45.0      & 50.6     & 50.9 \\
		Omniglot & 60.0     & 52.3      & 45.2     & 45.7 \\
		Aircraft & 53.1     & 49.0      & 36.0     & 37.2 \\
		Birds    & 68.8     & 62.2      & 60.7     & 61.2 \\
		Textures & 66.6     & 64.2      & 67.5     & 67.4 \\
		Quick Draw & 49.0   & 42.9      & 42.3     & 42.9 \\
		Fungi    & 39.7     & 34.0      & 30.1     & 31.3 \\
		VGG Flower & 85.3   & 80.1      & 70.7     & 71.0 \\
		Traffic Sign & 47.1 & 47.8      & 53.3     & 53.1 \\
		MSCOCO   & 41.0     & 35.0      & 45.2     & 45.6 \\ 
		MNIST    &          &           & 70.4     & 72.3 \\
		CIFAR10  &          &           & 65.2     & 66.7 \\
		CIFAR100 &          &           & 53.6     & 54.3 \\
		\hline
	\end{tabular}
\end{table}

\subsection{Clustering Analysis of Extracted Features in AZS Style}
\label{subsec:clustering_analyses}

Similar to the reasons described in the ANIL, the adaptation part cannot be 
removed in the classifiers. This means that the corresponding support sets cannot be 
removed in the classifiers of CNAPs.
Seeing this, the clustering performance of features, which are extracted by the feature extractors without adaptation, 
is analyzed in this section.
The PCA is used to visualize the extracted features for showing the better linear separability of CNAP-CMF features.

\begin{figure}[ht]
	\centering
	\subfigure[QucikDraw with CNAPs]{
		\includegraphics[scale=0.35]{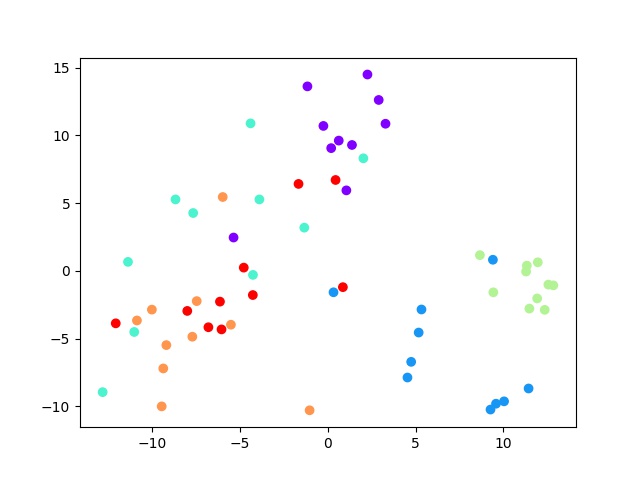}
		\label{subfigs:clustering_comparison_quick_CNAPs} 
	}
	\subfigure[QucikDraw with CNAP-CMF]{
		\includegraphics[scale=0.35]{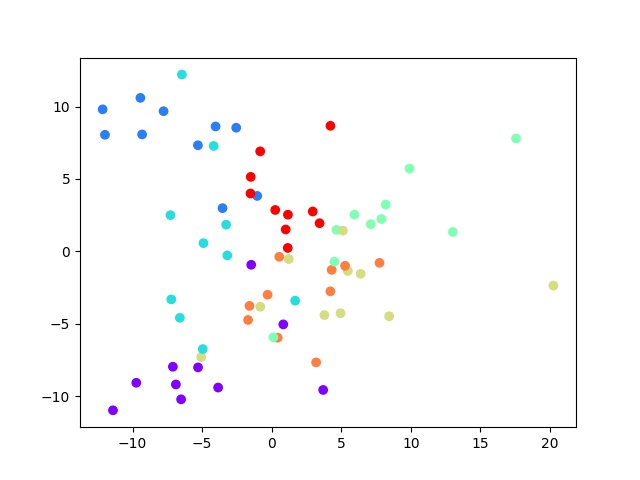}
		\label{subfigs:clustering_comparison_quick_CMFs}
	}
	\subfigure[Textures with CNAPs]{
		\includegraphics[scale=0.35]{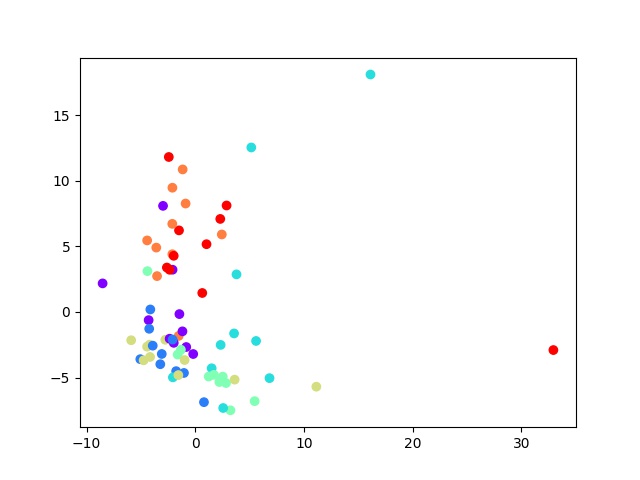}
		\label{subfigs:clustering_comparison_dtd_CNAPs} 
	}
	\subfigure[Textures with CNAP-CMF]{
		\includegraphics[scale=0.35]{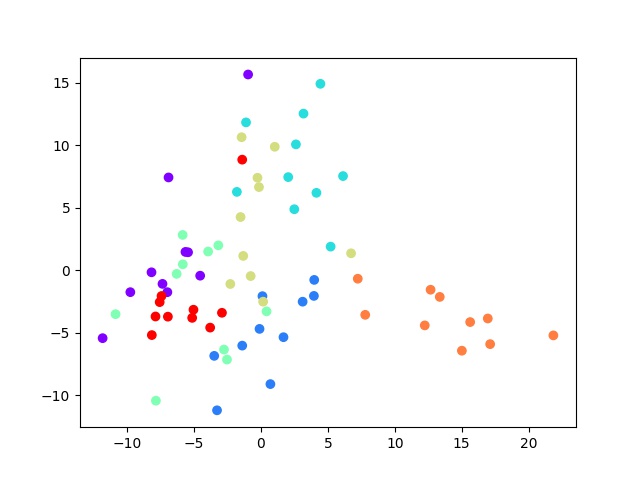}
		\label{subfigs:clustering_comparison_dtd_CMFs}
	}
	\caption{The clusterable property analyses of the extracted features with CNAPs and CMFs
		on QuickDraw and Textures datasets. The fixed support sets are randomly selected in 
	    ImageNet-1K.}
	\label{figs:clustering_analysis_features}
\end{figure}

The example figures shown in the right column of Fig. \ref{figs:clustering_analysis_features} are the 512-D features 
extracted by CNAPs in AZS style, which are linear inseparable, especially the features of Texture sub-dataset.
As a comparison, the features extracted by CNAP-CMF form obvious separable clusters, which are shown in the right column and 
indicate the more powerful feature extractor.

\begin{table}[ht]
	\centering
	\small
	\caption{Teh clusterable property comparison is analyzed by inner-class and inter-class Mahalanobis distance 
		of the extracted features between CNAPs and CNAP-CMF for AZS tasks.}
	\label{tab:clustering_analysis_inner_intra_dis}
	
	\begin{tabular}{l|c|c|c|c}
		\hline
		\multirow{2}{*}{Datasets} & \multicolumn{2}{|c|}{CNAPs} & \multicolumn{2}{|c}{CNAP-CMF} \\
		& Inner-class dist & Inter-class dist & Inner-class dist & Inter-class dist \\
		\hline
		Omniglot & 6.57 & 12.74 & 2.25 & 11.91 \\
		Aircraft & 4.78 & 9.23 & 1.51 & 8.73 \\
		Texture & 5.62 & 10.54 & 1.73 & 9.19 \\
		QucikDraw & 5.64 & 10.92 & 1.78 & 10.18 \\
		\hline
		
	\end{tabular}
\end{table}

For quantitative analysis, the inner-class and inter-class Mahalanobis distance between the extracted features are also 
given in Table \ref{tab:clustering_analysis_inner_intra_dis}. 
Firstly, PCA is used to reduce the dimensionality of the original 512-D feature embedding to 64-D
for reducing the effect of unimportant features because Mahalanobis distance is more sensitive to 
the noised features than Euclidean distance.
The Mahalanobis distance is used because it can reflect the distance along the data distribution \cite{2020simpleCNAPs}.

As given in Table \ref{tab:clustering_analysis_inner_intra_dis}, for example, on Omniglot, the inner-class distance of CNAP-CMF is 2.25, 
and the corresponding inner-class distance of CNAPs is 6.57. Simultaneously, the inter-class distance of CNAP-CMF is 11.91, while 
the corresponding inter-class distance of CNAPs is 12.74. Obviously, the much smaller inner-class distance and similar inter-class 
make the extracted features, without adaptation, of CNAP-CMF more linear separable than CNAPs.
Similarly, the clustering quantitative analysis on Aircraft, Texture, and QucikDraw also verifies the effectiveness of CNAP-CMF.

{\color{blue}
\subsection{Parameter Reduction and Inference Acceleration at The Test Stage}

The mean prior $\mathcal{R}^{C}_m$ generated by the mean encoder of CNAP-CMF is stable, and can be 
computed with any arbitrary support set. Thus,  
the mean prior $\mathcal{R}^{C}_m$ and the vectors for parameter shifting, 
$ (\gamma_w \otimes \mathcal{R}^{C}_m) $ and $ (\gamma_b \otimes \mathcal{R}^{C}_m) $, 
can be pre-computed in Eq. \ref{eq:adaption_resnet}.
Consequently, the two sub-networks, the mean encoder and the adaptation network, can be removed, 
which hence allows for the inference to be accelarated. 
The two sub-networks account for about 40\% parameter amount. 
For each sub-dataset of Meta-Dataset, the running time of CNAPs and CNAP-CMF 
for the sampled 600 tasks is computed. 
Table \ref{tab:light_test_stage} gives the parameter amount and inference time. 
The parameter amount of CNAPs is reduced from 21.3M to 12.7M, 
and the inference time is reduced from (3.5 + 20.7 + 12.3)s
to 12.3s.

\begin{table}[h]
	\centering
	\normalsize
	\caption{\color{blue} The parameter amount and
     the average running time for 600 tasks (for each sub-dataset) over all sub-datasets of Meta-Dataset.}
	\label{tab:light_test_stage}
	
	\begin{tabular}{l|cccc}                                                                                    
           \hline 
		     & \# Param (M) & Encoder (s) & Adaptation (s) & ResNet (s) \\
		\hline
		CNAPs       & 21.3    & 3.5        & 20.7                & 12.3       \\
		CNAP-CMF   & 12.7    & 0.0        & 0.0                 & 12.3       \\
           \hline 
	\end{tabular}
\end{table}
}

\section{Conclusion}
\label{sec:conclusion}

This paper investigated the role of feature reuse and the adaptation in the 
metric-based few-shot method, CNAPs. 
We designed two Almost Zero-Shot (AZS) experiments, AZS-I and AZS-II. 
In AZS-I, a support set is randomly chosen from each sub-dataset and 
used on all tasks within it; in AZS-II, 
a support set is randomly chosen from a sub-dataset but  
used on all tasks of all sub-dataset. Both AZS-I and AZS-II showed that 
the support sets were not necessary and the performance decrease of CNAPs without
the adaptation originates from the over-sensitiveness of the mean embeddings to the variation of
support sets. To deal with the over-sensitiveness, 
we proposed a Canonical Mean Filter that can map varying support sets 
to a canonical form. The CMF can allow for the adaptation to be 
stopped in CNAPs without the performance drop, and the stopped 
adaptation leads to 40.48\% parameter reduction in the test stage.
The analysis on the clustering property of extracted features by CNAP-CMF
showed that they were more separable than the features extracted by CNAPs, 
verifying that the feature reuse was more important.

\bibliography{egbib}

\begin{thebibliography}{10}
\expandafter\ifx\csname url\endcsname\relax
  \def\url#1{\texttt{#1}}\fi
\expandafter\ifx\csname urlprefix\endcsname\relax\def\urlprefix{URL }\fi
\expandafter\ifx\csname href\endcsname\relax
  \def\href#1#2{#2} \def\path#1{#1}\fi

\bibitem{NIPS2016MatchingNet}
O.~Vinyals, C.~Blundell, T.~Lillicrap, k.~kavukcuoglu, D.~Wierstra,
  \href{https://proceedings.neurips.cc/paper/2016/file/90e1357833654983612fb05e3ec9148c-Paper.pdf}{Matching
  networks for one shot learning}, in: D.~Lee, M.~Sugiyama, U.~Luxburg,
  I.~Guyon, R.~Garnett (Eds.), Advances in Neural Information Processing
  Systems, Vol.~29, Curran Associates, Inc., 2016.
\newline\urlprefix\url{https://proceedings.neurips.cc/paper/2016/file/90e1357833654983612fb05e3ec9148c-Paper.pdf}

\bibitem{finn17maml}
C.~Finn, P.~Abbeel, S.~Levine,
  \href{http://arxiv.org/abs/1703.03400}{Model-{A}gnostic {M}eta-{L}earning for
  {F}ast {A}daptation of {D}eep {N}etworks}, 2017.
\newline\urlprefix\url{http://arxiv.org/abs/1703.03400}

\bibitem{Raghu2020Rapid}
A.~Raghu, M.~Raghu, S.~Bengio, O.~Vinyals,
  \href{https://openreview.net/forum?id=rkgMkCEtPB}{Rapid learning or feature
  reuse? towards understanding the effectiveness of maml}, in: International
  Conference on Learning Representations, 2020.
\newline\urlprefix\url{https://openreview.net/forum?id=rkgMkCEtPB}

\bibitem{requeima2019cnaps}
J.~Requeima, J.~Gordon, J.~Bronskill, S.~Nowozin, R.~E. Turner, Fast and
  flexible multi-task classification using conditional neural adaptive
  processes, in: H.~Wallach, H.~Larochelle, A.~Beygelzimer, F.~d\'
  Alch\'{e}-Buc, E.~Fox, R.~Garnett (Eds.), Advances in Neural Information
  Processing Systems 32, Curran Associates, Inc., 2019, pp. 7957--7968.

\bibitem{Eleni2020MetaDataset}
E.~Triantafillou, T.~Zhu, V.~Dumoulin, P.~Lamblin, U.~Evci, K.~Xu, R.~Goroshin,
  C.~Gelada, K.~Swersky, P.-A. Manzagol, H.~Larochelle, Meta-dataset: A dataset
  of datasets for learning to learn from few examples, in: International
  Conference on Learning Representations, 2020.

\bibitem{Science2015Omniglot}
B.~M. Lake, R.~Salakhutdinov, J.~B. Tenenbaum,
  \href{https://science.sciencemag.org/content/350/6266/1332}{Human-level
  concept learning through probabilistic program induction}, Science 350~(6266)
  (2015) 1332--1338.
\newblock \href
  {http://arxiv.org/abs/https://science.sciencemag.org/content/350/6266/1332.full.pdf}
  {\path{arXiv:https://science.sciencemag.org/content/350/6266/1332.full.pdf}},
  \href {http://dx.doi.org/10.1126/science.aab3050}
  {\path{doi:10.1126/science.aab3050}}.
\newline\urlprefix\url{https://science.sciencemag.org/content/350/6266/1332}

\bibitem{Garnelo2020CNP}
G.~M, R.~D, M.~CJ, R.~T, S.~D, S.~M, T.~YW, R.~DJ, EslamiSM, Conditional neural
  processes, in: International Conference on Machine Learning, 2018.

\bibitem{Snell2017ProtoNets}
J.~Snell, K.~Swersky, R.~Zemel,
  \href{https://proceedings.neurips.cc/paper/2017/file/cb8da6767461f2812ae4290eac7cbc42-Paper.pdf}{Prototypical
  networks for few-shot learning}, in: I.~Guyon, U.~V. Luxburg, S.~Bengio,
  H.~Wallach, R.~Fergus, S.~Vishwanathan, R.~Garnett (Eds.), Advances in Neural
  Information Processing Systems, Vol.~30, Curran Associates, Inc., 2017.
\newline\urlprefix\url{https://proceedings.neurips.cc/paper/2017/file/cb8da6767461f2812ae4290eac7cbc42-Paper.pdf}

\bibitem{Biswas2018FirstorderMI}
A.~Biswas, S.~Agrawal, First-order meta-learned initialization for faster
  adaptation in deep reinforcement learning, 2018.

\bibitem{Nichol2018OnFM}
A.~Nichol, J.~Achiam, J.~Schulman, On first-order meta-learning algorithms,
  ArXiv abs/1803.02999.

\bibitem{WANG2019202}
D.~Wang, Y.~Cheng, M.~Yu, X.~Guo, T.~Zhang,
  \href{https://www.sciencedirect.com/science/article/pii/S0925231219305363}{A
  hybrid approach with optimization-based and metric-based meta-learner for
  few-shot learning}, Neurocomputing 349 (2019) 202--211.
\newblock \href
  {http://dx.doi.org/https://doi.org/10.1016/j.neucom.2019.03.085}
  {\path{doi:https://doi.org/10.1016/j.neucom.2019.03.085}}.
\newline\urlprefix\url{https://www.sciencedirect.com/science/article/pii/S0925231219305363}

\bibitem{garnelo2018neural}
M.~Garnelo, J.~Schwarz, D.~Rosenbaum, F.~Viola, D.~J. Rezende, S.~Eslami, Y.~W.
  Teh, Neural processes, arXiv preprint arXiv:1807.01622.

\bibitem{kim2019attentive}
H.~Kim, A.~Mnih, J.~Schwarz, M.~Garnelo, A.~Eslami, D.~Rosenbaum, O.~Vinyals,
  Y.~W. Teh, Attentive neural processes, arXiv preprint arXiv:1901.05761.

\bibitem{gordon2019convolutional}
J.~Gordon, W.~P. Bruinsma, A.~Y. Foong, J.~Requeima, Y.~Dubois, R.~E. Turner,
  Convolutional conditional neural processes, arXiv preprint arXiv:1910.13556.

\bibitem{foong2020convLNP}
A.~Y. Foong, W.~P. Bruinsma, J.~Gordon, Y.~Dubois, J.~Requeima, R.~E. Turner,
  Meta-learning stationary stochastic process prediction with convolutional
  neural processes, arXiv preprint arXiv:2007.01332.

\bibitem{kingma2013vae}
D.~P. Kingma, M.~Welling, Auto-encoding variational bayes, arXiv preprint
  arXiv:1312.6114.

\bibitem{2020DynamicConv}
Y.~Chen, X.~Dai, M.~Liu, D.~Chen, L.~Yuan, Z.~Liu,
  \href{http://arxiv.org/abs/1912.03458}{Dynamic convolution: Attention over
  convolution kernels}, in: Proceedings of the IEEE Conference on Computer
  Vision and Pattern Recognition, 2020.
\newline\urlprefix\url{http://arxiv.org/abs/1912.03458}

\bibitem{perez2018film}
E.~Perez, F.~Strub, H.~de~Vries, V.~Dumoulin, A.~C. Courville, Film: Visual
  reasoning with a general conditioning layer, in: AAAI, 2018.

\bibitem{He2015Deep}
K.~He, X.~Zhang, S.~Ren, et~al., Deep residual learning for image recognition,
  in: IEEE Conference on Computer Vision and Pattern Recognition, 2015, pp.
  770--778.

\bibitem{patacchiola2020bayesian}
M.~Patacchiola, J.~Turner, E.~J. Crowley, A.~Storkey, Bayesian meta-learning
  for the few-shot setting via deep kernels, in: Advances in Neural Information
  Processing Systems, 2020.

\bibitem{Robbins2020EMPIRICALBAYES}
H.~Robbins, \href{https://doi.org/10.1525/9780520313880-015}{AN EMPIRICAL BAYES
  APPROACH TO STATISTICS}, University of California Press, 2020, pp. 157--164.
\newblock \href {http://dx.doi.org/doi:10.1525/9780520313880-015}
  {\path{doi:doi:10.1525/9780520313880-015}}.
\newline\urlprefix\url{https://doi.org/10.1525/9780520313880-015}

\bibitem{2020Dynamic}
Y.~Chen, X.~Dai, M.~Liu, D.~Chen, Z.~Liu, Dynamic convolution: Attention over
  convolution kernels, in: 2020 IEEE/CVF Conference on Computer Vision and
  Pattern Recognition (CVPR), 2020.

\bibitem{Liu2020E3BM}
Y.~Liu, B.~Schiele, Q.~Sun, An ensemble of epoch-wise empirical bayes for
  few-shot learning, in: European Conference on Computer Vision (ECCV), 2020.

\bibitem{Kingma2014Adam}
D.~P. Kingma, J.~Ba, \href{http://arxiv.org/abs/1412.6980}{Adam: {A} method for
  stochastic optimization}, in: Y.~Bengio, Y.~LeCun (Eds.), 3rd International
  Conference on Learning Representations, {ICLR} 2015, San Diego, CA, USA, May
  7-9, 2015, Conference Track Proceedings, 2015.
\newline\urlprefix\url{http://arxiv.org/abs/1412.6980}

\bibitem{maji13fine-grained}
S.~Maji, J.~Kannala, E.~Rahtu, M.~Blaschko, A.~Vedaldi, Fine-grained visual
  classification of aircraft, Tech. rep. (2013).
\newblock \href {http://arxiv.org/abs/1306.5151} {\path{arXiv:1306.5151}}.

\bibitem{2020simpleCNAPs}
P.~Bateni, R.~Goyal, V.~Masrani, F.~Wood, L.~Sigal, Improved few-shot visual
  classification, in: Proceedings of the IEEE/CVF Conference on Computer Vision
  and Pattern Recognition (CVPR), 2020.

\end{thebibliography}

\clearpage
\appendix

\section{AZS-I Experiments}

\begin{table}[htb]
	\centering
	\normalsize
	\caption{Multi-task image classification results. CNAPs and CNAP-CMF are tested on 
		AZS-I and one-shot tasks: CNAPs-One, CNAPs-AZS, and CNAP-CMF-AZS-I.
	}
	\label{tab:AZS-I_experments}
	\begin{tabular}{l|ccc}
		\hline
		Datasets & CNAP-One & CNAP-AZS-I & CMF-AZS-I \\
		\hline
		ImageNet & 51.3     & 50.9       & 51.2 \\
		Omniglot & 88       & 87.2       & 87.6 \\
		Aircraft & 76.8     & 75.9       & 77.2 \\
		Birds    & 71.4     & 70.2       & 71.4 \\
		Textures & 62.5     & 63.1       & 63.0 \\
		Quick Draw & 71.9   & 71.5       & 70.5 \\
		Fungi    & 46       & 45.4       & 43.3 \\
		VGG Flower & 89.2   & 87.3       & 89.6 \\
		Traffic Sign & 60.1 & 58.5       & 62.0 \\
		MSCOCO   & 42       & 41.7       & 43.4 \\ 
		MNIST    & 88.6     & 87.1       & 89.1 \\
		CIFAR10  & 60       & 58.7       & 65.3 \\
		CIFAR100 & 48.1     & 47.6       & 50.9 \\
		\hline
	\end{tabular}
\end{table}

\section{Some Clustering Visualization}

\begin{figure*}[ht]
	\centering
	\subfigure[Aircraft]{
		\includegraphics[width=0.23 \textwidth]{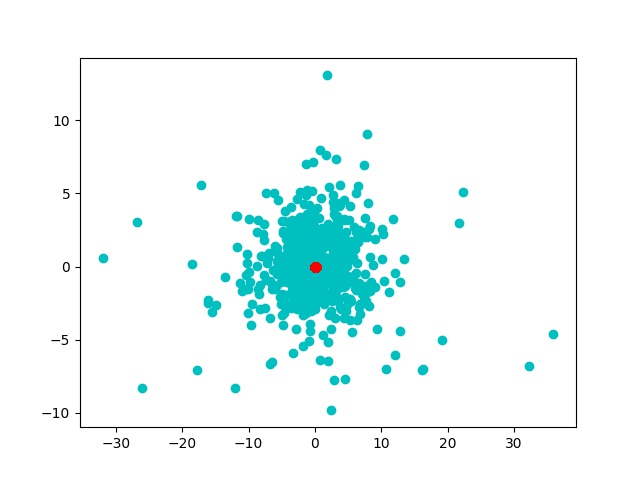}
		\label{subfig:clustering_aircraft}
	}
	\subfigure[ImageNet]{
		\includegraphics[width=0.23 \textwidth]{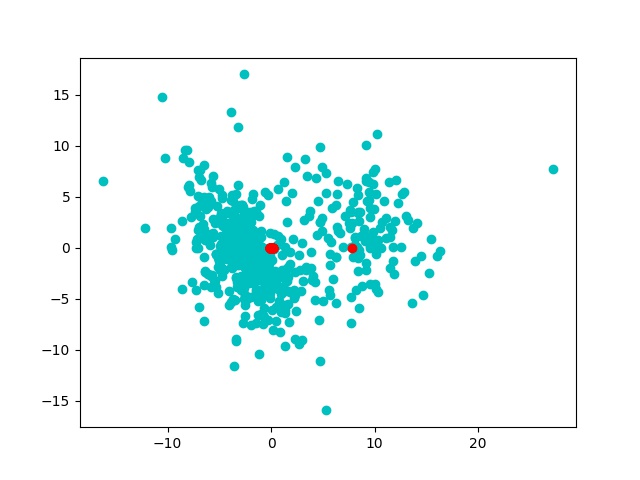}
		\label{subfig:clustering_ilsvrc_2012}
	}
	\subfigure[Omniglot]{
		\includegraphics[width=0.23 \textwidth]{clustering_omniglot.jpg}
		\label{subfig:clustering_omniglot}
	}
	\subfigure[Cu Birds]{
		\includegraphics[width=0.23 \textwidth]{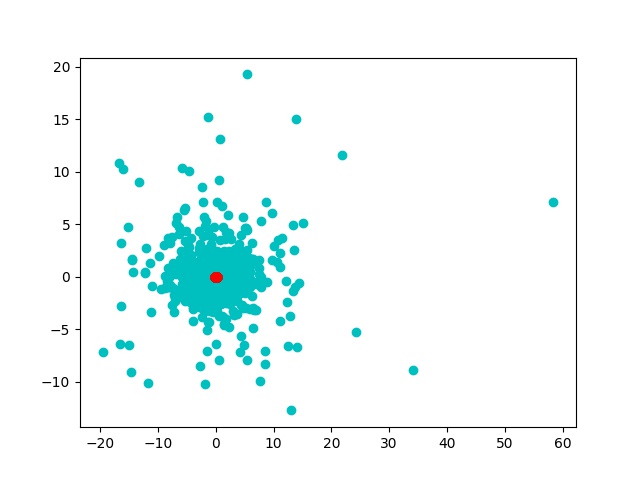}
		\label{subfig:clustering_cu_birds}
	}
	\subfigure[Dtd Texture]{
		\includegraphics[width=0.23 \textwidth]{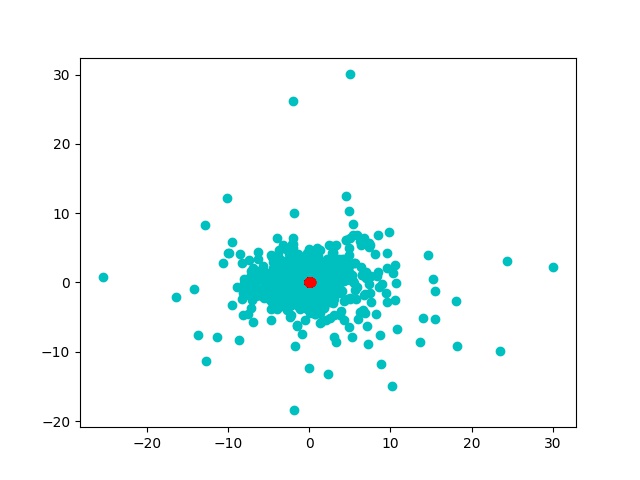}
		\label{subfig:clustering_dtd}
	}
	\subfigure[Fungi]{
		\includegraphics[width=0.23 \textwidth]{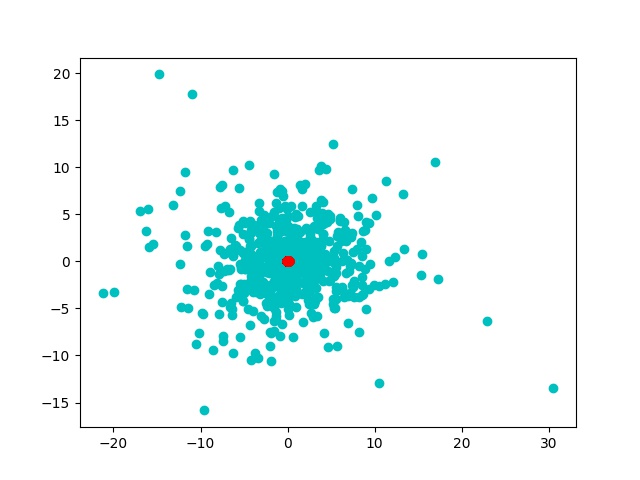}
		\label{subfig:clustering_fungi}
	}
	\subfigure[VGG Flower]{
		\includegraphics[width=0.23 \textwidth]{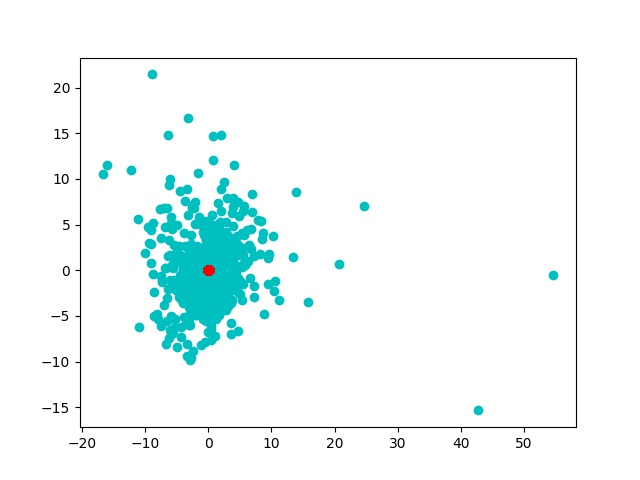}
		\label{subfig:clustering_vgg_flower}
	}
	\subfigure[QuickDraw]{
		\includegraphics[width=0.23 \textwidth]{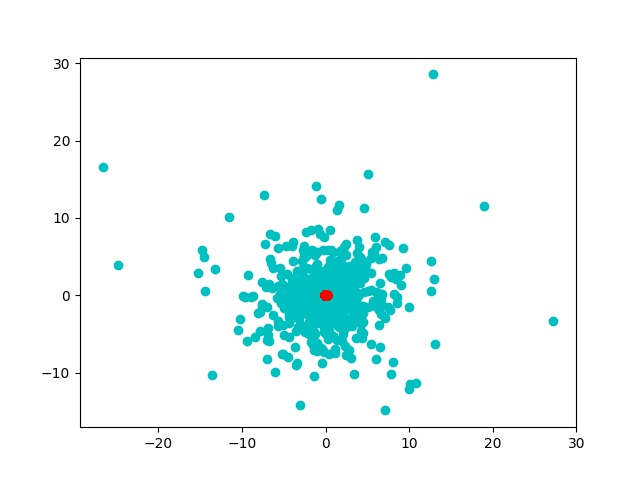}
		\label{subfig:clustering_quickdraw}
	}
	\subfigure[Traffic Sign]{
		\includegraphics[width=0.23 \textwidth]{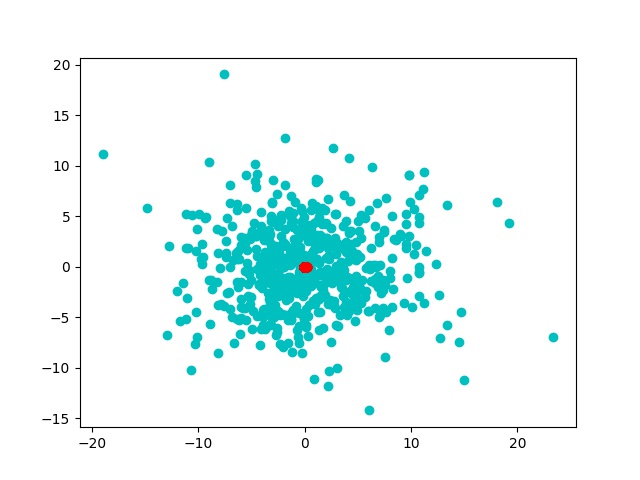}
		\label{subfig:clustering_traffic_sign}
	}
	\subfigure[MSCOCO]{
		\includegraphics[width=0.23 \textwidth]{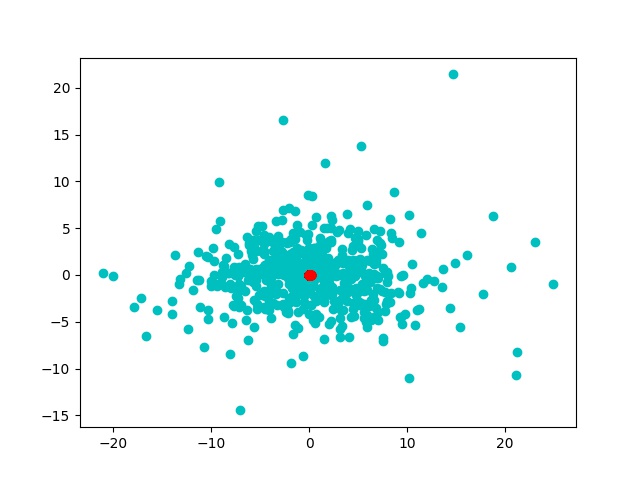}
		\label{subfig:clustering_mscoco}
	}
	\subfigure[MNIST]{
		\includegraphics[width=0.23 \textwidth]{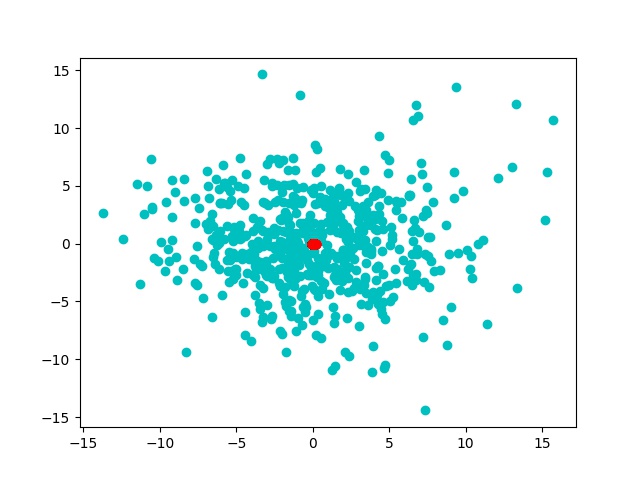}
		\label{subfig:clustering_mnist}
	}
	\subfigure[CIFAR-100]{
		\includegraphics[width=0.23 \textwidth]{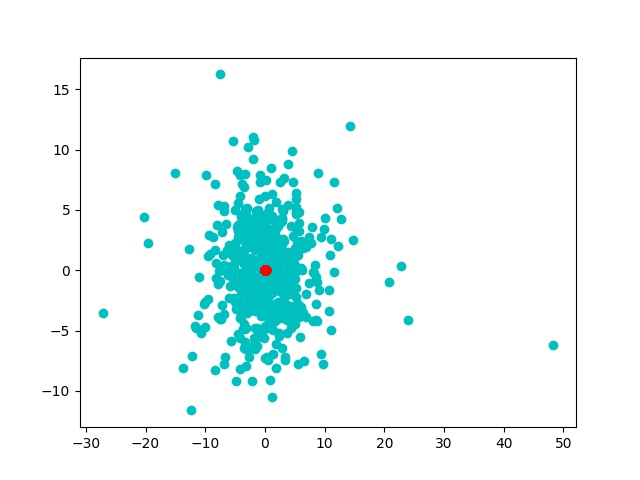}
		\label{subfig:clustering_cifar100}
	}
	\caption{Clustering Visualization of $\mathcal{R}^T_i$ (where $i \in \{1,2, \dots, N\}$, 
		and $N$ is the number of test image datasets in Meta-Datasets) with the model trained on Meta-Datasets. 
		Two dimensionality principal components are extracted by PCA.}
	\label{fig:clustering_visualization}
\end{figure*}

\begin{figure*}[ht]
	\centering
	\subfigure[Aircraft]{
		\includegraphics[width=0.3 \textwidth]{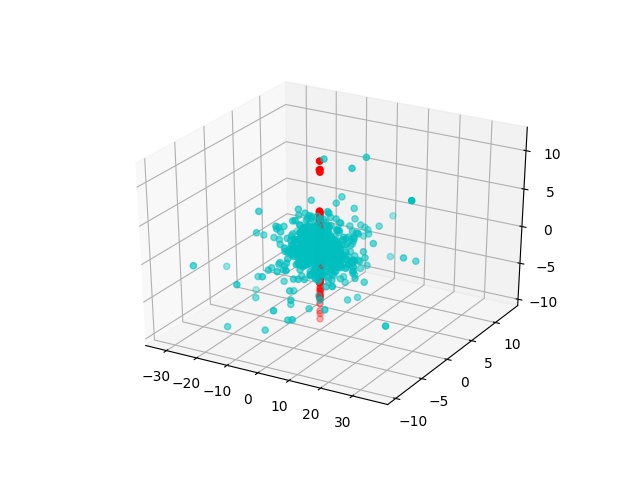}
		\label{subfig:clustering_aircraft_3D}
	}
	\subfigure[ImageNet]{
		\includegraphics[width=0.3 \textwidth]{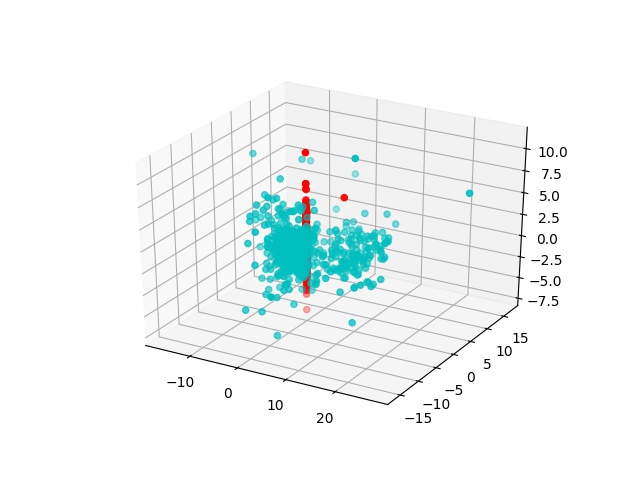}
		\label{subfig:clustering_ilsvrc_2012_3D}
	}
	\subfigure[Omniglot]{
		\includegraphics[width=0.3 \textwidth]{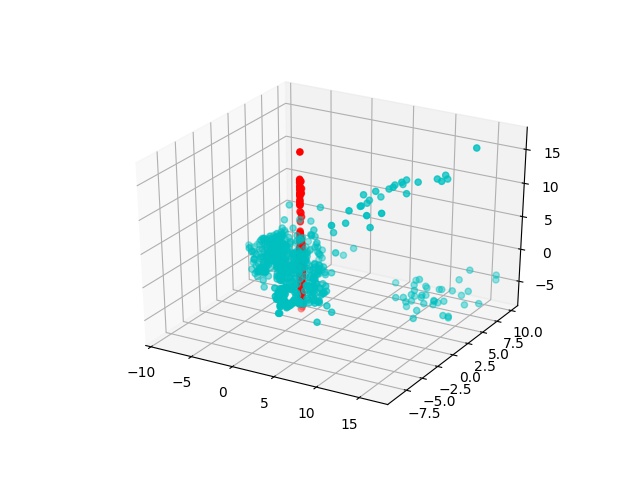}
		\label{subfig:clustering_omniglot_3D}
	}
	\subfigure[Cu Birds]{
		\includegraphics[width=0.3 \textwidth]{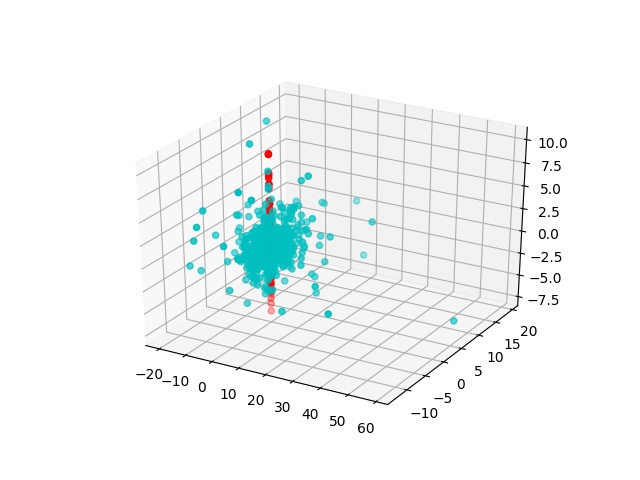}
		\label{subfig:clustering_cu_birds_3D}
	}
	\subfigure[Dtd Texture]{
		\includegraphics[width=0.3 \textwidth]{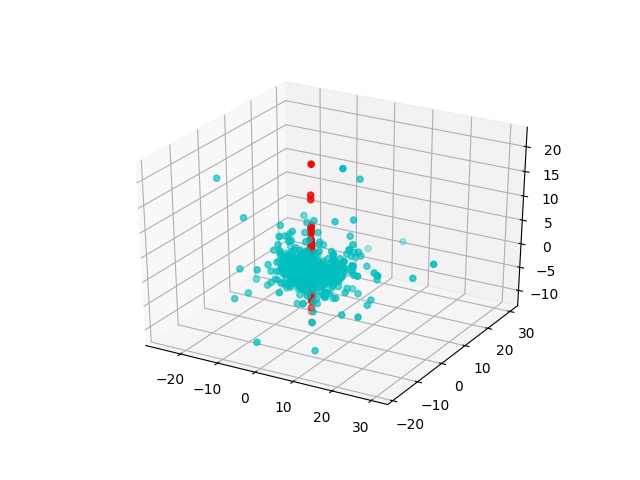}
		\label{subfig:clustering_dtd_3D}
	}
	\subfigure[Fungi]{
		\includegraphics[width=0.3 \textwidth]{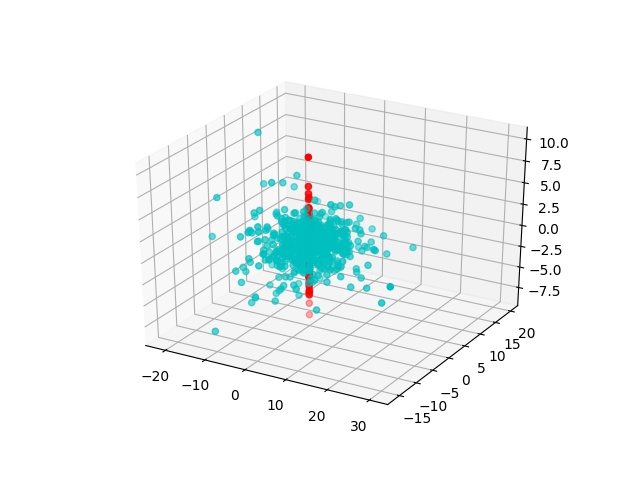}
		\label{subfig:clustering_fungi_3D}
	}
	\subfigure[VGG Flower]{
		\includegraphics[width=0.3 \textwidth]{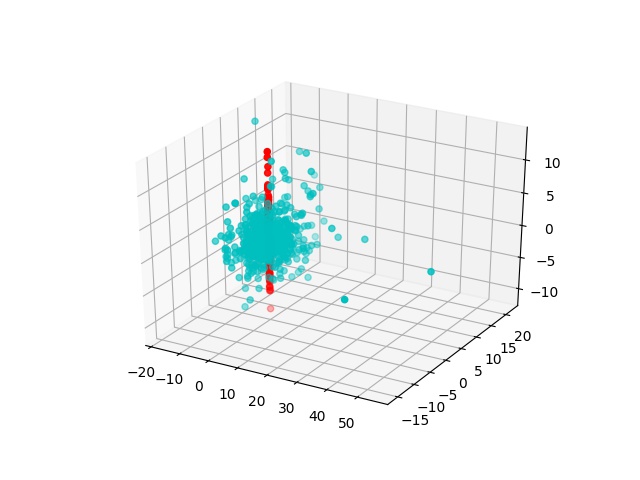}
		\label{subfig:clustering_vgg_flower_3D}
    }
	\subfigure[QuickDraw]{
		\includegraphics[width=0.3 \textwidth]{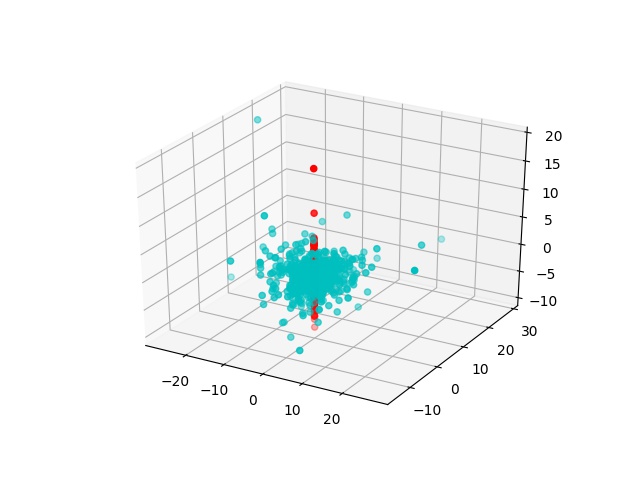}
		\label{subfig:clustering_quickdraw_3D}
	}
	\subfigure[Traffic Sign]{
		\includegraphics[width=0.3 \textwidth]{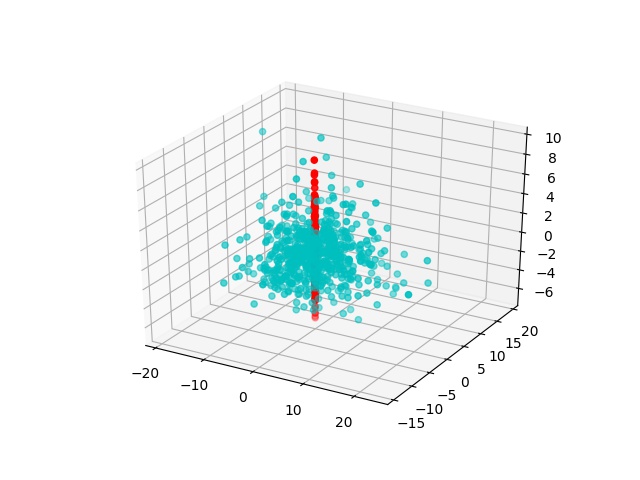}
		\label{subfig:clustering_traffic_sign_3D}
	}
	\subfigure[MSCOCO]{
		\includegraphics[width=0.3 \textwidth]{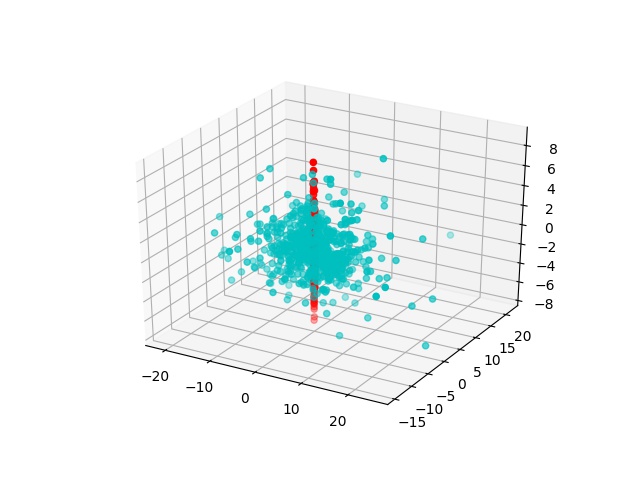}
		\label{subfig:clustering_mscoco_3D}
	}
	\subfigure[MNIST]{
		\includegraphics[width=0.3 \textwidth]{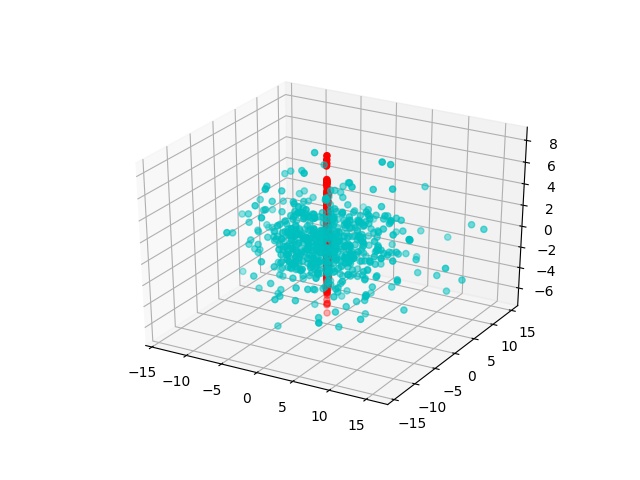}
		\label{subfig:clustering_mnist_3D}
	}
	\subfigure[CIFAR-100]{
		\includegraphics[width=0.3 \textwidth]{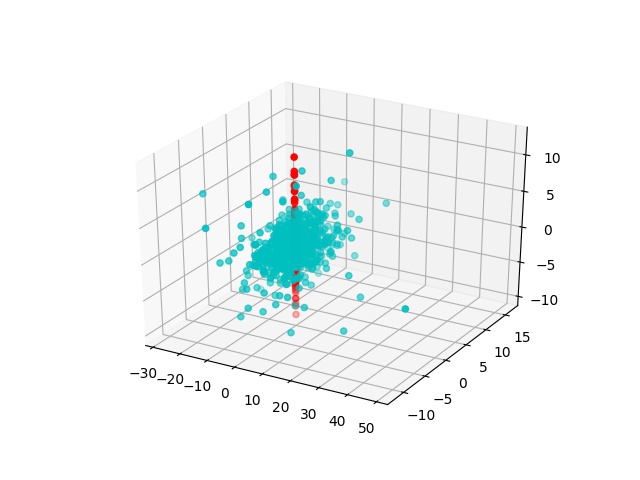}
		\label{subfig:clustering_cifar100_3D}
    }
	\caption{Clustering Visualization of $\mathcal{R}^T_i$ (where $i \in \{1,2, \dots, N\}$, 
	and $N$ is the number of test image datasets in Meta-Datasets) with the model trained on Meta-Datasets. 
	Three dimensionality principal components are extracted by PCA.}
	\label{fig:clustering_visualization_3D}
\end{figure*}


\section{Different Support Sets Selection Comparison}

Actually, for AZS-\uppercase\expandafter{\romannumeral2}, 
the unstable CNAPs perform more poor than the fixed support sets are selected from ImageNet. 
As shown in Table \ref{tab:each_support_selection}, the sub-datasets in the first row 
mean when the fixed support sets are selected in the corresponding sub-dataset.
The column under the sub-datasets gives the corresponding evaluated performance.
Obviously, except the performance in bold, 
which means the evaluated performance on the test sub-dataset when the support sets are 
selected from the corresponding sub-dataset, the evaluated performance drops a lot on other sub-datasets.

\begin{sidewaystable}[htb]
	\centering
	\small
	\caption{The performance of CNAPs-AZS-II is evaluated on Meta-Datasets when 
		the fixed support sets are selected from the sub-datasets except ImageNet.}
	\label{tab:each_support_selection}
	\resizebox{200mm}{70mm}{
	\begin{tabular}{l|cccccccccccc}
		Datasets & Omniglot & Aircraft & Birds & Textures & Quick Draw & Fungi & VGG Flower & Traffic Sign & MSCOCO & MNIST & CIFAR10 & CIFAR100  \\
		\hline
		\textbf{CNAPs-AZS-\uppercase\expandafter{\romannumeral2}} &&&&&&&&&&&& \\
		ImageNet     & 31.0  & 38.7  & 45.5 & 43.0 & 42.1 & 50.1 & 46.5 & 50.1 & 50.9 & 33.2 & 51.3 & 50.8 \\
		Omniglot     & \textbf{85.1}  & 67.6  & 55.6 & 55.4 & 78.5 & 57.7 & 65.9 & 60.0 & 56.1 & 84.2 & 55.3 & 52.5 \\
		Aircraft     & 51.6  & \textbf{74.6}  & 55.5 & 54.7 & 67.2 & 59.3 & 68.9 & 55.2 & 44.7 & 53.3 & 37.2 & 36.2 \\
		Birds        & 29.1  & 51.1  & \textbf{67.6} & 66.7 & 53.9 & 67.9 & 62.3 & 69.9 & 69.8 & 33.4 & 67.7 & 65.8 \\
		Textures     & 49.9  & 56.7  & 58.4 & \textbf{55.6} & 56.0 & 63.8 & 60.8 & 64.6 & 65.7 & 50.6 & 64.9 & 64.4 \\
		Quick Draw   & 60.4  & 60.9  & 46.3 & 44.8 & \textbf{70.6} & 54.5 & 61.4 & 57.9 & 51.7 & 63.5 & 51.9 & 50.8 \\
		Fungi        & 19.0  & 30.9  & 39.6 & 35.8 & 34.0 & \textbf{42.7} & 41.1 & 42.5 & 41.2 & 21.8 & 38.2 & 37.6 \\
		VGG Flower   & 63.7  & 73.6  & 74.5 & 70.5 & 82.7 & 81.8 & \textbf{83.2} & 83.4 & 80.8 & 70.7 & 79.6 & 78.7 \\
		Traffic Sign & 56.9  & 47.4  & 48.4 & 47.7 & 53.6 & 49.7 & 53.7 & \textbf{53.0} & 54.5 & 56.4 & 56.8 & 58.0 \\
		MSCOCO       & 29.5  & 34.0  & 39.5 & 36.7 & 38.3 & 43.6 & 40.9 & 47.5 & \textbf{47.1} & 32.1 & 47.5 & 47.7 \\ 
		MNIST        & 90.6  & 86.4  & 74.9 & 74.6 & 87.9 & 76.2 & 83.6 & 78.2 & 78.6 & \textbf{90.5} & 79.2 & 79.3 \\
		CIFAR10      & 45.2  & 50.7  & 53.7 & 52.1 & 56.6 & 59.0 & 54.2 & 62.2 & 65.8 & 51.1 & \textbf{67.8} & 67.0 \\
		CIFAR100     & 29.8  & 33.4  & 38.1 & 34.3 & 40.0 & 44.0 & 40.1 & 49.7 & 52.2 & 32.5 & 53.4 & \textbf{54.1} \\
		\hline
		\textbf{CNAP-CMF-AZS-\uppercase\expandafter{\romannumeral2}} &&&&&&&&&&&& \\
		ImageNet     & 50.8  & 51.5  & 51.0 & 50.6 & 50.7 & 50.6 & 51.4 & 50.4 & 50.5 & 51.2 & 50.4 & 50.5 \\
		Omniglot     & \textbf{88.0}  & 86.8  & 87.8 & 88.0 & 88.2 & 88.5 & 88.0 & 87.6 & 88.0 & 87.6 & 87.9 & 87.9 \\
		Aircraft     & 73.4  & \textbf{76.6}  & 73.6 & 72.9 & 73.1 & 73.0 & 73.0 & 73.1 & 73.7 & 73.4 & 74.2 & 73.1 \\
		Birds        & 69.2  & 69.6  & \textbf{71.3} & 70.0 & 70.4 & 69.9 & 69.9 & 70.1 & 69.1 & 70.6 & 70.2 & 70.0 \\
		Textures     & 59.1  & 60.8  & 59.7 & \textbf{62.9} & 59.3 & 60.0 & 60.0 & 59.8 & 59.6 & 60.2 & 59.9 & 59.5 \\
		Quick Draw   & 70.4  & 70.0  & 69.7 & 70.1 & \textbf{70.6} & 70.4 & 69.5 & 70.0 & 69.7 & 70.3 & 70.7 & 69.9 \\
		Fungi        & 43.4  & 43.4  & 43.4 & 42.1 & 43.6 & \textbf{44.2} & 43.6 & 43.8 & 44.3 & 43.5 & 43.0 & 44.1 \\
		VGG Flower   & 87.5  & 87.9  & 87.9 & 88.5 & 87.5 & 87.9 & \textbf{88.2} & 88.0 & 87.3 & 88.4 & 88.1 & 88.1 \\
		Traffic Sign & 60.0  & 61.8  & 61.1 & 60.1 & 60.2 & 61.5 & 60.7 & \textbf{62.1} & 61.2 & 61.6 & 61.7 & 61.4 \\
		MSCOCO       & 45.4  & 43.9  & 45.5 & 44.0 & 45.3 & 44.7 & 44.4 & 45.6 & \textbf{46.9} & 44.4 & 44.2 & 44.1 \\ 
		MNIST        & 89.0  & 88.1  & 88.9 & 88.5 & 88.5 & 88.9 & 88.5 & 88.5 & 88.9 & \textbf{89.5} & 89.2 & 88.8 \\
		CIFAR10      & 62.5  & 62.7  & 62.7 & 62.6 & 62.5 & 62.9 & 63.2 & 62.2 & 62.8 & 63.6 & \textbf{63.8} & 63.6 \\
		CIFAR100     & 49.7  & 48.8  & 49.4 & 47.1 & 47.8 & 47.5 & 48.3 & 47.9 & 47.7 & 48.1 & 47.8 & \textbf{49.3} \\
	\end{tabular}}
\end{sidewaystable}

\end{document}